\documentclass{article}
\usepackage{conference,times}

\usepackage{amsmath,amsfonts,bm}

\def\eqref#1{equation~\ref{#1}}

\def\1{\bm{1}}

\DeclareMathAlphabet{\mathsfit}{\encodingdefault}{\sfdefault}{m}{sl}
\SetMathAlphabet{\mathsfit}{bold}{\encodingdefault}{\sfdefault}{bx}{n}

\usepackage{hyperref}
\usepackage{url}
\usepackage{graphicx}
\usepackage{multirow}
\usepackage{multicol}
\usepackage{booktabs}
\usepackage{xcolor}
\usepackage{bbm}

\title{VC-OPD: Learning to See What Matters via Visual Cue On-Policy Self-Distillation}

\author{
\makebox[\textwidth][c]{\begin{tabular}{c}
{\small\bfseries
Qinxuan Wang\textsuperscript{1,2,3,\ensuremath{\dagger}}
\enspace
Zheyu Zhang\textsuperscript{3,\ensuremath{\dagger}}
\enspace
Yijing Yang\textsuperscript{3,\ensuremath{\dagger}}
\enspace
Han Shu\textsuperscript{3,*}
\enspace
Chuang Wang\textsuperscript{1,2,*}
} \\[0.25em]
{\small\bfseries
Peipei Yang\textsuperscript{1,2}
\quad
Shuo Tang\textsuperscript{1,2}
\quad
Jie Hu\textsuperscript{3}
\quad
Shiming Xiang\textsuperscript{1,2}
} \\[0.8em]
{\normalfont
\textsuperscript{1} MAIS, Institute of Automation,
Chinese Academy of Sciences
} \\
{\normalfont
\textsuperscript{2} School of Artificial Intelligence,
University of Chinese Academy of Sciences
} \\
{\normalfont
\textsuperscript{3} Huawei Technologies Ltd.
} \\[0.5em]
{\normalfont\footnotesize
\textsuperscript{\ensuremath{\dagger}}
Equal contribution.
} \\
{\normalfont\footnotesize
\textsuperscript{*}
Corresponding authors: Han Shu and Chuang Wang.
} \\
{\normalfont\footnotesize
\texttt{han.shu@huawei.com} \quad \texttt{wangchuang@ia.ac.cn}
} \\
{\normalfont\footnotesize
Work done during an internship at Huawei Noah's Ark Lab (Qinxuan Wang \texttt{wqinxuan4@gmail.com}).
} 
\end{tabular}}
}

\iclrfinalcopy
\begin{document}

\maketitle
\lhead{VC-OPD: Learning to See What Matters via Visual Cue On-Policy Self-Distillation}

\begin{abstract}
Existing visual On-Policy Distillation (OPD) methods typically use cropped or zoomed-in image regions as teacher-side privileged information to improve the fine-grained visual perception of small Multimodal Large Language Models (MLLMs). However, such local pixel-level information is insufficient for complex visual reasoning that requires understanding global structures, spatial relationships, and high-level semantics.
To address this limitation, we propose Visual Cue On-Policy Self-Distillation (VC-OPD), which extends teacher-side privileged information from local image regions to question-conditioned visual cues. Specifically, we leverage large advanced MLLMs to extract the key visual evidence required to solve a given question, including key visual details, structural relationships, and high-level semantics. A multi-stage quality filtering mechanism is further introduced to retain reliable visual cues as privileged information. During training, these cues are provided only to the teacher model, enabling it to deliver more informative token-level supervision along student-generated trajectories. The student is thereby encouraged to internalize the ability to identify and reason over critical visual evidence directly from the original image, without requiring any additional information at inference time. Since large advanced MLLMs are used only for visual cue generation, VC-OPD is applicable to both open-weight and proprietary black-box models.
Experiments on STEM and general visual reasoning benchmarks show that VC-OPD consistently improves visual reasoning performance across different model scales and outperforms a range of reinforcement learning and on-policy distillation baselines. The 4B model trained with VC-OPD surpasses many baselines with 8B or more parameters, demonstrating the effectiveness of question-conditioned visual cues as privileged information for capability transfer.

\end{abstract}

\section{Introduction}

In recent years, Multimodal Large Language Models (MLLMs) have achieved remarkable progress in visual understanding, visual question answering, and complex multimodal reasoning tasks~\citep{Llava-onevision,Qwen3vl2025}. As visual reasoning tasks increasingly extend to challenging scenarios in mathematics, science, and engineering (STEM), models are required not only to recognize visual content, but also to understand the spatial, geometric, and structural relationships among visual elements and perform multi-step reasoning conditioned on the given question~\citep{mathvista,mathverse,mmmu,xue2022multi}. However, small-scale MLLMs still lag considerably behind larger models on complex visual understanding and reasoning tasks. Therefore, effectively transferring stronger visual understanding and reasoning capabilities to smaller models while maintaining low inference costs remains an important challenge in building efficient MLLMs.

Knowledge Distillation (KD) provides an effective approach for capability transfer~\citep{hinton2015}. Recently, On-Policy Distillation (OPD) has alleviated the distribution shift inherent in conventional offline distillation by providing teacher supervision on trajectories generated by the student model itself~\citep{GKD2024}. Building on this paradigm, On-Policy Self-Distillation (OPSD) provides the teacher model with privileged information that is inaccessible to the student at inference time, enabling the teacher to produce more informative learning signals along the student's on-policy trajectories~\citep{OPSD2026}. In visual settings, using cropped and zoomed-in local images as teacher-side privileged information can substantially enhance the fine-grained visual perception capabilities of small MLLMs~\citep{Vision-OPD2026}.
However, such privileged information primarily supplements local pixel-level details. Complex STEM visual reasoning requires models to further identify which visual information is relevant to the current question and understand the spatial and semantic relationships among different visual elements. Therefore, relying solely on local visual enhancement is insufficient to fully support the demands of complex visual reasoning.

More importantly, we find that the effectiveness of privileged information depends not only on the improvement it brings to the teacher's problem-solving capability, but also on the transferability of the resulting supervision to the student. Privileged information that directly reveals the solution can make the teacher substantially stronger, yet may introduce informational shortcuts that the student cannot recover from its original multimodal input. This suggests that teacher capability alone is insufficient for effective on-policy distillation and that privileged information should also be task-relevant and recoverable from the student's input, allowing the knowledge exposed to the teacher to be effectively internalized by the student.

\begin{figure}[t]
\centering
\centerline{\includegraphics[width=\linewidth]{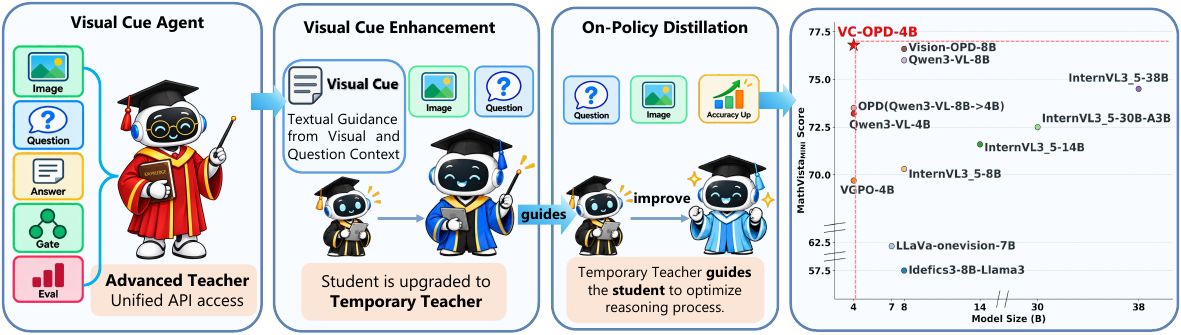}}
\caption{Overview and performance of VC-OPD. A visual cue agent translates a large MLLM's multimodal understanding into question-conditioned visual cues. Used as teacher-side privileged information in OPD, visual cues enable compact MLLMs to achieve competitive performance against substantially larger models.
}
\label{fig-compare-others}
\end{figure}

Meanwhile, transferring capabilities from stronger large advanced MLLMs is also constrained by the teacher access regime. Conventional white-box knowledge distillation typically relies on the teacher model's output distributions, hidden representations, or other internal information to provide dense supervision signals. However, for proprietary large models that are accessible only through APIs, such internal information is generally unavailable, and distillation must instead rely on observable model outputs, resulting in black-box knowledge distillation. These differences in access privileges make it difficult for existing methods to fully exploit both open-weight and proprietary large models within a unified framework.

Motivated by the above observations, this paper proposes Visual Cue On-Policy Self-Distillation (VC-OPD), which introduces question-conditioned visual cues as privileged context accessible to the teacher model, as shown in Fig.~\ref{fig-compare-others}. To enable the unified use of large advanced MLLMs under different access regimes, VC-OPD incorporates a Visual Cue Agent that is compatible with both open-weight models and proprietary models accessible only through black-box interfaces. Given an image and a question, the agent leverages a large advanced MLLM to extract the key visual information required to solve the current task, including salient objects and their attributes, spatial and structural relationships, and high-level visual semantics relevant to reasoning. In addition, a series of quality-gating mechanisms is employed to select samples near the current capability boundary of the small MLLM and to filter out candidate visual cues that are unreliable or difficult to utilize effectively. Through this process, the stronger visual understanding and reasoning capabilities of large advanced MLLMs are transformed into compact, task-relevant, and validated visual cues, which are subsequently transferred to small MLLMs through on-policy self-distillation.

The main contributions of this work are summarized as follows:

\begin{itemize}
\item We extend teacher-side privileged information to question-conditioned, high-level \textbf{visual cues}, enabling the teacher to access more task-relevant visual evidence and better capture the spatial, structural, and semantic relationships among visual elements. These visual cues are also highly transferable, as the evidence they encode can be recovered from the student's original multimodal input, allowing the student to internalize the corresponding visual understanding and reasoning capabilities.

\item We propose \textbf{VC-OPD} and introduce an access-agnostic \textbf{Visual Cue Agent}, enabling both open-weight and proprietary large advanced MLLMs to provide transferable visual understanding and reasoning knowledge through a unified visual-cue interface.

\item Extensive experiments on multiple STEM and general visual reasoning benchmarks demonstrate that VC-OPD substantially improves the visual reasoning capabilities of small MLLMs, enabling smaller models to match or outperform larger-scale baselines across multiple benchmarks.
\end{itemize}

\section{Related Work}

\paragraph{Off-Policy Distillation.}

Traditional knowledge distillation (KD) typically optimizes student models using pre-collected, fixed training trajectories~\citep{hinton2015}. Kim et al.~\citep{kim2016sequence} extend this idea from token-level prediction to complete sequences, using teacher-generated sequences as supervision for the student.
Later, Hsieh et al.~\citep{hsieh2023distilling} further leverages answers and natural-language reasoning rationales pre-generated by a powerful LLM as static supervision signals for training smaller models.

\paragraph{On-Policy Distillation.}

Generalized Knowledge Distillation (GKD)~\citep{GKD2024} performs token-level teacher distillation on trajectories generated by the student itself, thereby mitigating the distribution shift between training and inference states and providing dense supervision over the states actually visited by the student.
Building on this idea, On-Policy Self-Distillation (OPSD)~\citep{OPSD2026} further allows the teacher and student to share the same model parameters, creating a capability gap solely through asymmetric access to information.
On-Policy Context Distillation (OPCD)~\citep{OPCD2026} extends this paradigm to more general context-distillation settings, where additional context, such as prior problem-solving experience and optimized system prompts, is provided as teacher-side information and the knowledge encoded in this context is internalized into a student that operates without such additional context.
Rethinking OPD~\citep{li2026rethinking} further shows that additional Reinforcement Learning (RL) post-training of the teacher can endow it with new capabilities that the student has not yet acquired, substantially improving the effectiveness of subsequent knowledge transfer through OPD.
These works suggest that the teacher's advantage in on-policy distillation does not necessarily arise from a larger model scale, but can also be constructed through additional information or context available during training.

\paragraph{On-Policy Distillation for MLLMs.}

As the effectiveness of on-policy distillation for LLM reasoning has been increasingly validated, recent studies have begun to extend this paradigm to MLLMs.
VOLD~\citep{VOLD2026} explores using a text-only LLM as the teacher and transferring its reasoning capabilities to a vision-language model through on-policy distillation.
More recent work has further investigated how to construct privileged teachers with stronger access to visual information within MLLMs.
Vision-OPD~\citep{Vision-OPD2026} provides the teacher with locally cropped and enlarged images containing critical visual evidence, while the student continues to observe the full image. In this way, the model internalizes the teacher's stronger fine-grained perception under the zoomed-in view into the student policy.
OPD-V~\citep{OPD-V2026} constructs a Positive Teacher that observes a zoomed-in image and a Negative Teacher that observes a masked image from the perspective of modality balance. The logit margin between the two teachers is then used to define a Modality-Balance Trust Region, such that distillation is applied only to on-policy tokens that reflect meaningful contributions from visual information.
Visual Attribution Distillation (VAD)~\citep{VAD2026} compares the same teacher under evidence-present and evidence-removed views, thereby providing more targeted visual supervision.
Visual-OPSD~\citep{Visual-OPSD2026} uses intermediate visual images generated by a unified multimodal model as privileged context for the teacher during training. By exploiting the cross-modal conditioning discrepancy between teacher and student, it distills the reasoning knowledge encoded along the visual generation pathway into a student that does not require intermediate visual image generation.
However, privileged information constructed purely from pixel-level visual signals, such as locally zoomed-in images, may be insufficient for capturing the spatial, structural, and semantic relationships among different visual elements.

\section{Methodology}

\subsection{Overview}

\begin{figure}[h]
\centering
\centerline{\includegraphics[width=\linewidth]{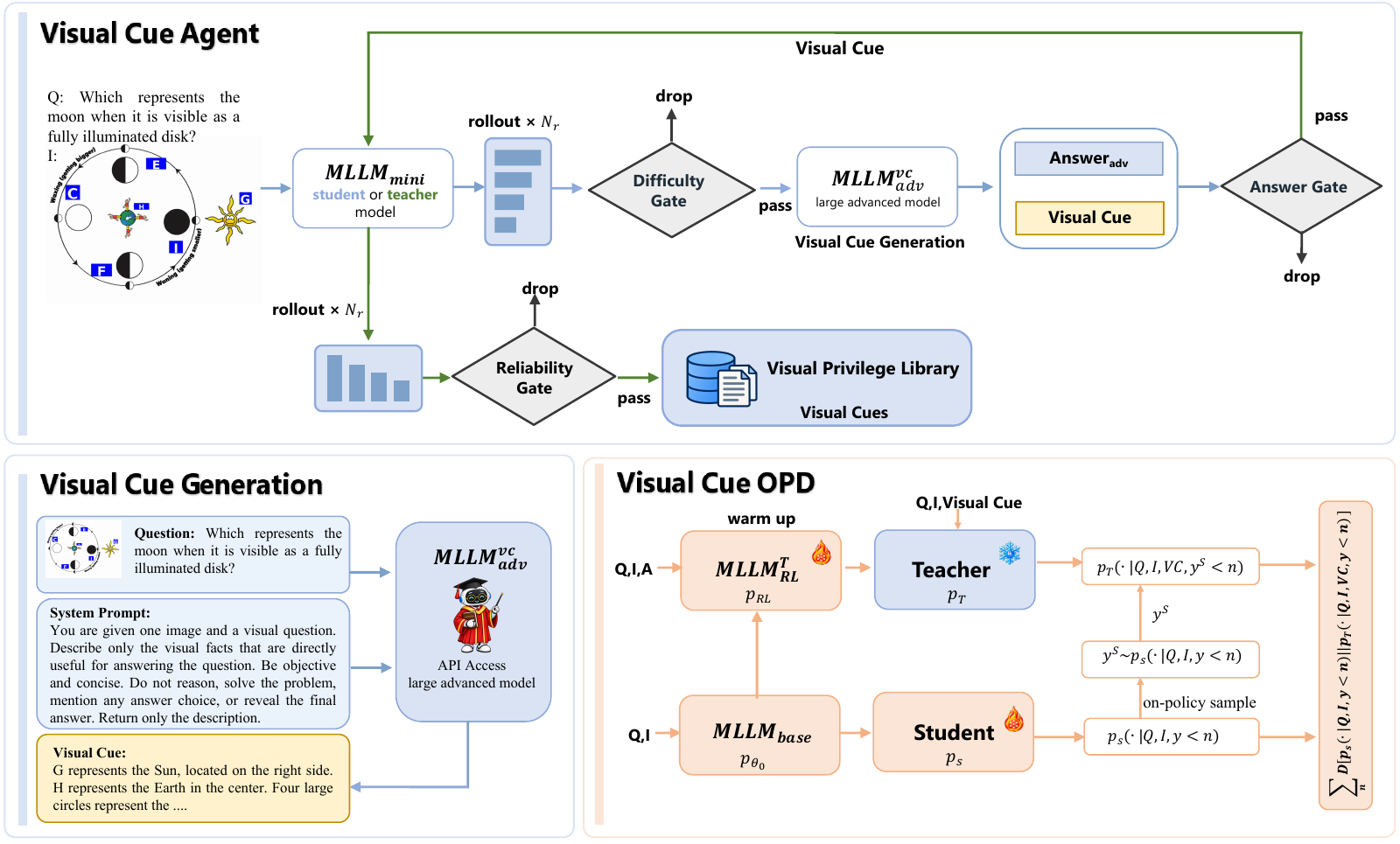}}
\caption{Overview of VC-OPD. The Visual Cue Agent generates privileged visual information for the small model. It uses a large MLLM to produce task-relevant visual cues and a small MLLM to assess their quality. Visual Cue OPD then internalizes the agent’s multimodal understanding into the small model through OPD.
}
\label{fig-overview}
\end{figure}

The proposed VC-OPD uses question-conditioned visual cues as teacher-side privileged information to transfer stronger visual understanding and reasoning capabilities from large advanced MLLMs to compact MLLMs. As shown in Fig.~\ref{fig-overview}, VC-OPD consists of two stages. First, the Visual Cue Agent leverages a large advanced MLLM to extract essential visual evidence from the original image, transforming the MLLM's visual understanding into compact visual cues that capture key objects, relationships, and semantic information. A gating mechanism then filters unreliable samples and cues to construct the Visual Privilege Library. Based on this library, Visual Cue OPD transfers the knowledge encoded in visual cues through a privileged teacher model to the student MLLM. The student performs on-policy rollouts using only the original image and question, while the frozen teacher receives the corresponding visual cue at the same student-generated states. By minimizing the discrepancy between their predictive distributions, the student gradually internalizes the ability to identify and utilize question-conditioned visual evidence directly from the original image. Since visual cues are only used during training and the large advanced MLLM is required only for cue generation, VC-OPD is applicable to both open models and black-box models.

\subsection{Visual Cue Agent}
\label{sec:visual_cue_agent}
The Visual Cue Agent converts the stronger visual understanding and reasoning capabilities of a large advanced MLLM into compact, question-conditioned privileged visual information.
Given a training sample, inspired by rejection sampling in LLM post-training~\citep{touvron2023llama}, the agent progressively screens the generated visual cues through three quality gates: a Difficulty Gate that identifies samples near the current capability boundary of the student model, an Answer Gate that verifies the correctness of the large advanced MLLM's visual understanding, and a Reliability Gate that evaluates whether the generated visual cue can reliably support correct problem solving.
Only visual cues that pass all three gates are retained in the Visual Privilege Library for subsequent on-policy self-distillation.

\paragraph{Difficulty Gate.}
We first evaluate the difficulty of each sample $(I,q,a)$ using multiple rollouts from the initial student policy $\pi_{\theta_0}$.
The Difficulty Gate is defined as
\begin{equation}
G_{\mathrm{diff}}
=
\mathbbm{1}\!\left[
\tau_{\mathrm{d}}^{-}
\leq
\sum_{j=1}^{N_r}\mathcal{E}(y_j^{(0)},a)
\leq
\tau_{\mathrm{d}}^{+}
\right],
\quad
y_j^{(0)}\sim\pi_{\theta_0}(\cdot\mid I,q),
\label{eq:difficulty_gate}
\end{equation}
where $I$, $q$, and $a$ represent the image, question, and answer, respectively. $\mathcal{E}(y,a)\in\{0,1\}$ denotes the answer evaluation function, which returns $1$ if the generated response $y$ is judged correct with respect to the reference answer $a$, and $0$ otherwise.
$N_r$ denotes the number of rollouts, while $\tau_{\mathrm{d}}^{-}$ and $\tau_{\mathrm{d}}^{+}$ specify the lower and upper bounds on the number of correct rollouts, respectively.
This gate filters out samples that are either too easy or too difficult, retaining those near the student's current capability boundary.

\paragraph{Answer Gate.}
For each sample passing the Difficulty Gate, we query a large advanced MLLM
($M_{\mathrm{adv}}^{\mathrm{VC}}$) separately to obtain a predicted answer
$\hat{a}_{\mathrm{adv}}$ and a question-conditioned visual cue $v$:
\begin{equation}
\hat{a}_{\mathrm{adv}}
=
M_{\mathrm{adv}}^{\mathrm{VC}}(I,q;\mathcal{P}_{\mathrm{Ans}}),
\quad
v
=
M_{\mathrm{adv}}^{\mathrm{VC}}(I,q;\mathcal{P}_{\mathrm{VC}}),
\label{eq:visual_cue_generation}
\end{equation}
where $\mathcal{P}_{\mathrm{Ans}}$ prompts the model to answer the question,
while $\mathcal{P}_{\mathrm{VC}}$ instructs it to extract only visual evidence relevant to the current question, such as object attributes, textual information, and spatial relationships.
The final answer is explicitly excluded from $v$ to prevent answer leakage.

The predicted answer is used solely for quality verification through the Answer Gate:
\begin{equation}
G_{\mathrm{ans}}
=
\mathbbm{1}\!\left[
\mathcal{E}(\hat{a}_{\mathrm{adv}},a)=1
\right].
\label{eq:answer_gate}
\end{equation}
We retain the corresponding visual cue $v$ only when $\hat{a}_{\mathrm{adv}}$ is correct, thereby filtering cues that may originate from erroneous visual understanding.

\paragraph{Reliability Gate.}
Although the Answer Gate verifies that the large advanced MLLM can correctly answer the question, it does not guarantee that the generated visual cue is useful for reasoning by a smaller MLLM. Therefore we provide the candidate visual cue \(v\) to the teacher policy \(\pi_T\) and evaluate its utility through multiple rollouts. The Reliability Gate is defined as
\begin{equation}
G_{\mathrm{rel}}
=
\mathbbm{1}\!\left[
\sum_{j=1}^{N_r}
\mathcal{E}(y_j^{(v)},a)
\geq
\tau_{\mathrm{rel}}
\right],
\quad
y_j^{(v)}
\sim
\pi_T(\cdot\mid I,q,v),
\label{eq:reliability_gate}
\end{equation}
where $\tau_{\mathrm{rel}}$ denotes the minimum number of correct rollouts required for a candidate cue to pass the gate.
This gate retains only visual cues that can consistently support correct problem solving by the teacher model.
Finally, the visual cues that pass all three quality gates form the Visual Privilege Library:
\begin{equation}
\mathcal{B}_{\mathrm{VC}}
=
\left\{
(I,q,a,v)
\;\middle|\;
G_{\mathrm{diff}}\times
G_{\mathrm{ans}}\times
G_{\mathrm{rel}}
=1
\right\}.
\label{eq:visual_privilege_library}
\end{equation}
The resulting library contains validated, question-conditioned visual cues that are provided exclusively to the teacher during the subsequent Visual Cue OPD stage. These cues are designed to be transferable, as the visual evidence they encode remains accessible from the student's original multimodal input, allowing the student to internalize the corresponding visual understanding and reasoning capabilities through distillation.

\subsection{Visual Cue OPD}
\label{sec:on_policy_self_distillation}
The Visual Cue Agent converts the visual understanding and reasoning capabilities of large advanced MLLMs into reliable visual cues.
We further employ on-policy self-distillation to transfer the visual understanding encoded in these cues to the student model.
Notably, large advanced MLLMs are used only for visual cue generation, while token-level distillation is performed between the teacher and student models.
Therefore, VC-OPD requires no access to the internal representations of large advanced MLLMs and can leverage both open-weight and proprietary black-box models through the same visual-cue interface.

\paragraph{Teacher Warm-up.}

To obtain a teacher model capable of providing stable and high-quality token-level supervision, we start from the base MLLM policy $\pi_{\theta_0}$ and perform a reasoning warm-up on the original visual reasoning training set $\mathcal{D}_{\mathrm{train}}$ using Group Relative Policy Optimization (GRPO), yielding the teacher policy $\pi_T$:
\begin{equation}
    \pi_T
    =
    \operatorname{GRPO}
    \left(
        \pi_{\theta_0};
        \mathcal{D}_{\mathrm{train}}
    \right).
\end{equation}
The objective is to strengthen the teacher's fundamental visual reasoning capability, enabling it to effectively exploit additional privileged visual information during the subsequent distillation stage. After the GRPO warm-up, we fix the teacher parameters and keep the teacher frozen throughout the entire OPD process. The student model is initialized from the same base policy $\pi_{\theta_0}$, ensuring that the teacher and student share an identical model architecture and output space.

\paragraph{Visual Cue Privileged Teacher.}

For each sample $(I,q,a,v)\in\mathcal{B}_{\mathrm{VC}}$ from the Visual Privilege Library, the student model performs an on-policy rollout under its current policy $\pi_S$ using only the original image $I$ and question $q$.
At the $n$-th position of the student-generated trajectory $y^S$, the student and frozen teacher are conditioned on the same student-generated prefix $y^S_{<n}$, while the teacher additionally receives the visual cue $v$.
Their corresponding next-token predictive distributions are
\begin{equation}
p_S^{(n)}=\pi_S(\cdot\mid I,q,y^S_{<n}), \qquad p_T^{(n)}=\pi_T(\cdot\mid I,q,v,y^S_{<n}).
\label{eq:student_teacher_distributions}
\end{equation}
Because the teacher and student always predict the next token from the same student-generated state, the discrepancy between their predictive distributions reflects the additional visual information provided by the visual cue for the current decision.
The teacher leverages the visual cue to form a more informative predictive distribution, while the student learns to extract and utilize question-conditioned visual evidence directly from the original image by matching this distribution.

\paragraph{Loss function.}
To transfer privileged visual information to the student on its own state distribution, we minimize the token-level Jensen--Shannon divergence (JSD) between the student and privileged teacher distributions along student-generated trajectories:
\begin{equation}
\mathcal{L}_{\mathrm{VC\text{-}OPD}}=\mathbb{E}_{(I,q,a,v)\sim\mathcal{B}_{\mathrm{VC}},\,y^S\sim\pi_S(\cdot\mid I,q)}\left[\frac{1}{L}\sum_{n=1}^{L}D_{\mathrm{JS}}\!\left(p_S^{(n)}\,\|\,p_T^{(n)}\right)\right].
\label{eq:vc_opd_objective}
\end{equation}
where $y^S$ is an on-policy trajectory sampled from the current student policy, $L$ denotes its sequence length, and $D_{\mathrm{JS}}(\cdot \|\ \cdot)$ is the Jensen--Shannon divergence. During optimization, the teacher remains frozen and only the student parameters are updated; the visual cue affects the student solely through the teacher's predictive distribution and is never directly provided to the student.

\section{Experiments}

\paragraph{Datasets.}
ViRL39K~\citep{ViRL39K} is used as the training dataset. We evaluate our method on a diverse suite of benchmarks covering both STEM reasoning and general visual reasoning, including MathVerse, MathVista, and MMMU-Pro, among others. Further details on the training and evaluation datasets are provided in Appendices~\ref{sec-Training-Dataset} and~\ref{sec-Benchmarks}.

Detailed hyperparameter settings are provided in Appendix~\ref{sec-Hyperparameter}. Ablation studies on the GRPO warm-up stage are presented in Appendix~\ref{sec-exp-warm-up}.

\subsection{Comparison with advanced methods}

\begin{table*}[t]
\centering
\caption{Comparison with open-weight MLLMs and representative post-training methods on STEM and general visual reasoning benchmarks. For post-training methods, all methods use Qwen3-VL-Instruct as the base model and are evaluated at three model scales: 2B, 4B, and 8B. Bold and underlined numbers indicate the best and second-best results within each model scale, respectively. The subscript ``$_{\mathrm{m}}$'' denotes the MINI version. For MathVerse$_{\mathrm{m}}$, the superscripts ``$\mathrm{v}$'' and ``$\mathrm{t}$'' denote the Vision-only and Text-dominant settings, respectively.
}
\setlength{\tabcolsep}{0.7pt}
\renewcommand{\arraystretch}{1.08}

\resizebox{\textwidth}{!}{\begin{tabular}{c|cccccc|c|ccc|c}
\toprule
& \multicolumn{7}{c|}{STEM Visual Reasoning Benchmarks}
& \multicolumn{4}{c}{General Visual Reasoning Benchmarks} \\
\midrule
Models
& MathVerse$_{\mathrm{m}}^{\mathrm{v}}$
& MathVerse$_{\mathrm{m}}^{\mathrm{t}}$
& DynaMath
& WeMath
& MathVista$_{\mathrm{m}}$
& MathVision$_{\mathrm{m}}$
& \textbf{\, Avg.\, }
& MMMU$_{\mathrm{pro}}^{\mathrm{v}}$
& MMMU$_{\mathrm{pro}}^{\mathrm{std}}$
& HallBen
& \textbf{\, Avg.\, } \\
\midrule

\multicolumn{12}{c}{\textit{Open-Weight Model Baselines}} \\
\midrule

LLaVA-OneVision-7B
& 17.01 & 39.21 & 34.49 & 17.62 & 62.30 & 18.42
& 31.51 & 14.45 & 30.52 & 57.31 & 34.09 \\

Idefics3-8B-Llama3
& 14.97 & 25.76 & 29.74 & 11.62 & 57.50 & 12.83
& 25.40 & 13.64 & 27.86 & 61.09 & 34.20 \\

InternVL3.5-8B
& 26.78 & 48.22 & 57.64 & 36.57 & 70.30 & 33.55
& 45.51 & 31.27 & 40.98 & 68.98 & 47.08 \\

InternVL3.5-14B
& 32.36 & 56.09 & 60.36 & 41.05 & 71.60 & 36.84
& 49.72 & 33.53 & 44.39 & 68.14 & 48.69 \\

InternVL3.5-30B-A3B
& 35.03 & 57.87 & 62.42 & 38.76 & 72.50 & 33.88
& 50.08 & 30.06 & 44.28 & 69.40 & 47.91 \\

InternVL3.5-38B
& 37.18 & 58.12 & 63.67 & 46.86 & 74.50 & 42.43
& 53.80 & 37.34 & 46.24 & 69.72 & 51.10 \\

\midrule
\multicolumn{12}{c}{\textit{Post-Training Method Baselines}} \\
\midrule
Qwen3-VL-2B (base)
& 34.14 & 51.14 & 53.05 & 30.48 & 62.70 & \textbf{29.28}
& 43.46 & 27.92 & 35.38 & \underline{68.98} & 44.09 \\

ReMax \citep{Remax2023}
& \underline{41.88} & \underline{57.23} & 53.27 & 34.10 & 64.80 & 28.29
& \underline{46.60} & \underline{31.04} & 36.18 & 68.56 & \underline{45.26} \\

GRPO \citep{GRPO2024}
& 35.66 & 53.68 & \underline{54.19} & \underline{35.71} & 62.40 & 28.29
& 44.99 & 30.98 & 35.43 & 67.19 & 44.54 \\

OPSD \citep{OPSD2026}
& 36.17 & 49.75 & 52.38 & 29.90 & 65.00 & \underline{28.95}
& 43.69 & 26.76 & 35.66 & 67.19 & 43.21 \\

VGPO \citep{VGPO2026}
& 37.94 & \textbf{58.12} & 51.68 & 34.19 & 59.40 & 28.29
& 44.94 & 29.65 & 33.18 & 65.93 & 42.92 \\

Vision-OPD \citep{Vision-OPD2026}
& 31.22 & 45.43 & 51.16 & 27.24 & 62.40 & 25.00
& 40.41 & 25.55 & 34.34 & 66.25 & 42.04 \\

\textbf{VC-OPD (Ours)}
& \textbf{45.05} & \textbf{58.12} & 54.13 & \textbf{38.19} & \textbf{67.00} & 26.32
& \textbf{48.14} & \textbf{31.73} & \textbf{37.63} & \textbf{69.93} & \textbf{46.43} \\

\midrule

Qwen3-VL-4B (base)
& 30.58 & 53.93 & 64.51 & 49.05 & 73.20 & 37.50
& 51.46 & 30.58 & 35.72 & 70.66 & 45.65 \\

ReMax \citep{Remax2023}
& 30.46 & 57.36 & 64.83 & 46.67 & 74.10 & 42.43
& 52.64 & 30.98 & 35.66 & 69.72 & 45.45 \\

GRPO \citep{GRPO2024}
& 32.36 & 57.49 & 63.59 & \underline{50.29} & 74.60 & \textbf{45.07}
& 53.90 & 30.12 & 35.49 & 64.35 & 43.32 \\

OPSD \citep{OPSD2026}
& \underline{53.05} & \underline{67.01} & 64.13 & 45.81 & \underline{75.20} & 38.82
& \underline{57.33} & 38.90 & \textbf{53.18} & 65.62 & 52.57 \\

VGPO \citep{VGPO2026}
& 31.98 & 48.35 & 62.36 & \underline{50.29} & 69.70 & 39.14
& 50.30 & 31.73 & 36.18 & 69.61 & 45.84 \\

Vision-OPD \citep{Vision-OPD2026}
& 48.60 & 66.88 & 64.79 & 46.67 & 74.00 & 36.18
& 56.19 & \underline{41.85} & 46.99 & 70.98 & \underline{53.27} \\

\textbf{VC-OPD (Ours)}
& \textbf{59.39} & \textbf{72.08} & \textbf{65.89} & \textbf{51.62}
& \textbf{76.80} & \underline{44.08}
& \textbf{61.64} & \textbf{45.49} & \underline{49.71} & \textbf{73.19} & \textbf{56.13} \\

\midrule

Qwen3-VL-8B (base)
& 36.55 & 67.26 & 67.37 & 53.90 & 76.60 & \underline{46.38}
& 58.01 & 42.43 & 41.27 & \underline{74.45} & 52.72 \\

ReMax \citep{Remax2023}
& 39.97 & \underline{71.07} & 66.93 & \textbf{59.71} & 77.30 & 43.09
& 59.68 & 42.08 & 42.43 & 73.50 & 52.67 \\

GRPO \citep{GRPO2024}
& 37.44 & 66.24 & 67.01 & 55.14 & 76.00 & 43.09
& 57.49 & 37.51 & 40.17 & 72.98 & 50.22 \\

OPSD \citep{OPSD2026}
& \underline{57.61} & 70.81 & 66.77 & 53.33 & 76.40 & 42.43
& \underline{61.23} & \underline{49.88} & 52.83 & 74.34 & \underline{59.02} \\

VGPO \citep{VGPO2026}
& 37.31 & 56.22 & 64.25 & 54.38 & \underline{77.90} & 39.80
& 54.98 & 37.11 & 39.83 & 71.50 & 49.48 \\

Vision-OPD \citep{Vision-OPD2026}
& 54.95 & 70.30 & \textbf{69.72} & 50.19 & 76.00 & 44.08
& 60.87 & 48.84 & \underline{52.89} & 73.40 & 58.38 \\

\textbf{VC-OPD (Ours)}
& \textbf{61.55} & \textbf{73.10} & \underline{68.82} & 55.33 & 76.80 & 45.07
& \textbf{63.44} & \textbf{52.49} & \textbf{53.76} & \textbf{74.87} & \textbf{60.37} \\

\bottomrule
\end{tabular}}
\label{tab-compare-others}
\end{table*}

To evaluate the effectiveness of VC-OPD, we first compare it with several mainstream open-weight MLLMs, including LLaVA-OneVision-7B~\citep{Llava-onevision}, Idefics3-8B-Llama3~\citep{Idefics3}, different scales of InternVL3.5~\citep{Internvl3.5}, and the Qwen3-VL~\citep{Qwen3vl2025} base models. We conduct a unified evaluation on STEM and general visual reasoning benchmarks using VLMEvalKit \citep{Vlmevalkit}, with the results presented in Table~\ref{tab-compare-others} and Fig.~\ref{fig-compare-others}. Despite being applied to relatively small models, VC-OPD enables them to achieve performance comparable to or even better than substantially larger open-weight models. For example, Qwen3-VL-2B post-trained with VC-OPD achieves a higher average score on the STEM benchmarks than InternVL3.5-8B, while attaining comparable performance on general visual reasoning tasks. Furthermore, VC-OPD-4B surpasses the substantially larger InternVL3.5-38B and Qwen3-VL-8B base model in average performance. These results demonstrate that VC-OPD can effectively enhance the visual reasoning capabilities of smaller models and narrow the performance gap between them and substantially larger models.

In addition, we compare VC-OPD with several representative post-training methods, including OPSD~\citep{OPSD2026}, GRPO~\citep{GRPO2024}, ReMax~\citep{Remax2023}, VGPO~\citep{VGPO2026}, and Vision-OPD~\citep{Vision-OPD2026}. For a fair comparison, all post-training methods adopt Qwen3-VL-Instruct as the base model and are evaluated at three model scales: 2B, 4B, and 8B. The results are summarized in Table~\ref{tab-compare-others}.
Across all three model scales, VC-OPD consistently improves upon the corresponding Qwen3-VL base models and achieves the best average performance among the post-training methods at each scale. Taking Qwen3-VL-4B as an example, VC-OPD improves the average score on the STEM benchmarks from 51.46 to 61.64, a gain of 10.18 percentage points. On the general visual reasoning benchmarks, the average score increases from 45.65 to 56.13, yielding a gain of 10.48 percentage points. Similar improvements are consistently observed in the 2B and 8B settings. These results indicate that the benefits of VC-OPD are not confined to STEM domains but generalize well to broader visual reasoning tasks. Moreover, the consistent improvements across model scales demonstrate the strong scalability of VC-OPD.

Compared with advanced post-training methods, VC-OPD provides more targeted and transferable supervision for visual reasoning. ReMax~\citep{Remax2023} and GRPO~\citep{GRPO2024} rely on sparse sequence-level rewards, offering limited fine-grained supervision for intermediate reasoning. OPSD~\citep{OPSD2026} uses ground-truth answers as privileged information, which may introduce answer shortcuts that are difficult to recover from the original image. VGPO~\citep{VGPO2026} enhances visual activation but lacks explicit guidance on critical visual content and its relations, while Vision-OPD~\citep{Vision-OPD2026} improves local perception through cropped or enlarged regions but may miss global structures and high-level semantics. In contrast, VC-OPD provides question-conditioned visual cues that describe key visual evidence while excluding final answers, making the privileged information both task-relevant and recoverable from the original image. 
This advantage is reflected in the results: under the 4B setting, VC-OPD achieves 61.64 and 56.13 on the STEM and general visual reasoning benchmarks, outperforming the strongest baselines, OPSD (57.33) and Vision-OPD (53.27), by 4.31 and 2.86 points, respectively. Moreover, VC-OPD maintains the highest average performance across the 2B, 4B, and 8B settings, demonstrating consistent gains and scalability across diverse visual reasoning tasks.

\subsection{Visualization of performance improvement}
This experiment demonstrates the effectiveness of visual cues in enhancing visual reasoning from two perspectives: the teacher model itself and the student model after OPD training.

\begin{figure}[t]
\centering
\centerline{\includegraphics[width=\linewidth]{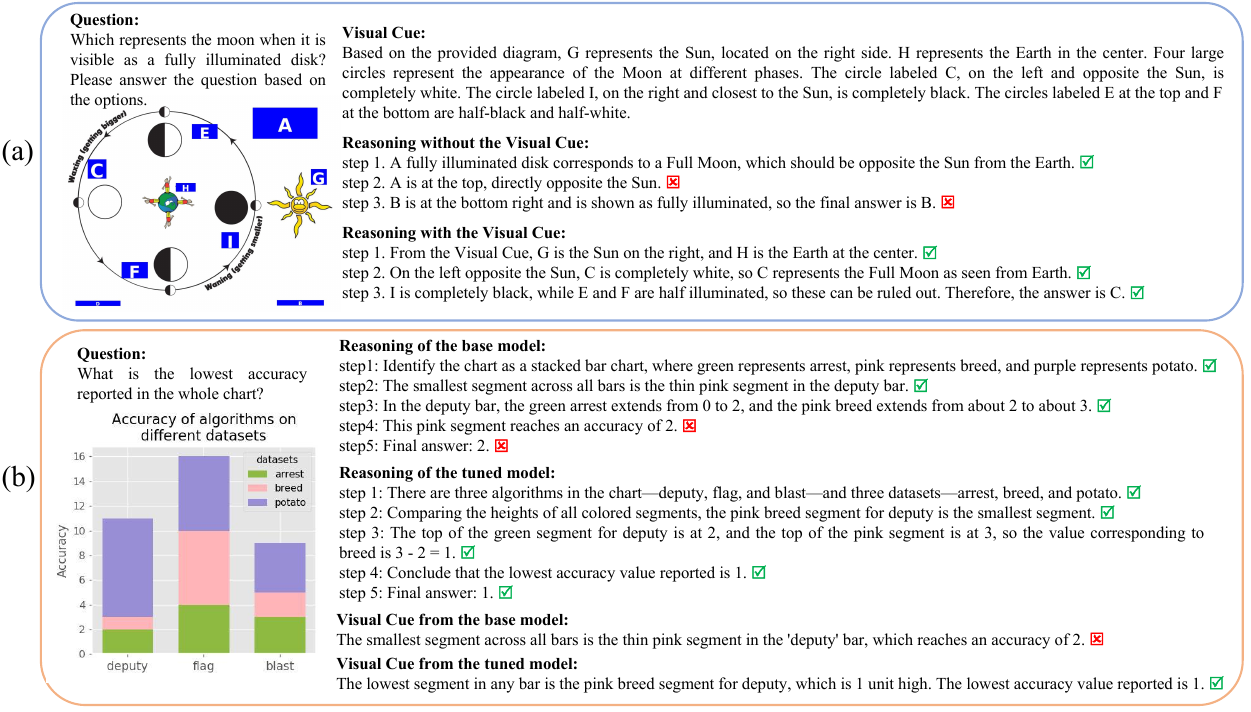}}
\caption{Visualization of the improvements achieved by VC-OPD. (a) Performance improvement of the teacher model with and without visual cues. (b) Performance improvement of the student model before and after training.
}
\label{fig-visual-cap}
\end{figure}

As shown in Fig.~\ref{fig-visual-cap}(a), without visual cues, the teacher model misidentifies the spatial relationships among celestial objects in an astrophysics problem. By providing accurate task-relevant information about visual attributes and spatial relationships, visual cues enable the teacher to correctly interpret the image and answer the question.

Fig.~\ref{fig-visual-cap}(b) illustrates the benefit of VC-OPD for the student model. Before training, the base model confuses the cumulative height of a stacked bar with the value of an individual segment. Specifically, the pink ``breed'' segment extends from \(y=2\) to \(y=3\) and thus represents a value of 1 rather than its absolute vertical coordinate. After training, the model correctly reasons over the segment height and recovers the correct value.

These examples show that visual cues enhance the teacher's visual perception and reasoning, while OPD transfers these capabilities to the student, enabling it to better reason over critical visual evidence without visual cues at inference time. More qualitative examples are provided in Appendix~\ref{sec-display-improve}.

\subsection{Ablation experiments on different forms of privileged information}

To investigate the effects of different forms of privileged information on On-Policy Distillation, we compare four types of teacher-side privileged information: reasoning process (Reasoning), ground-truth answer (Answer), generic image description (Caption), and the proposed question-conditioned visual cue (Visual Cue). All settings use the same student, warm-up teacher, and training configuration, differing only in the additional information accessible to the teacher.

The results in Table \ref{tab-privileged-ablation} show that the teacher's problem-solving capability with privileged information does not directly determine the final distillation performance. Although Answer and Reasoning both increase the teacher's Teacher Score to above 90, they do not yield the best average downstream performance. In contrast, Visual Cue achieves substantially lower Teacher Scores, ranging from 72.68 to 82.32, while obtaining the highest average downstream performance. This suggests that a stronger privileged teacher does not necessarily lead to more effective knowledge transfer; the transferability of the privileged information itself is equally important.

Answer and Reasoning provide relatively direct solution information. However, such information is unavailable in the student's input at inference time and may therefore create informational shortcuts that are difficult for the student to recover from the original image. Since Caption is generated without question conditioning, it often contains redundant or task-irrelevant information that may obscure critical evidence, while potentially omitting fine-grained details or relationships essential to the current question. In contrast, Visual Cue retains only question-conditioned information about key objects, attributes, and spatial and structural relationships, providing the teacher with more targeted visual evidence without directly revealing the final answer.

These results suggest that effective privileged information should not merely maximize the teacher's immediate problem-solving performance, but should also balance task relevance and recoverability. While strengthening the teacher, Visual Cue provides critical visual evidence that can, in principle, be learned and recovered by the student from the original image. This property may facilitate more effective teacher–student supervision, consistent with its stronger downstream distillation performance.

\begin{table*}[t]
\centering
\setlength{\tabcolsep}{2pt}
\renewcommand{\arraystretch}{1.15}
\label{tab-}
\caption{Ablation results on different forms of privileged information. Student denotes the base model. Teacher Score denotes the teacher model’s evaluation score on the training set when conditioned on the corresponding privileged information. Bold numbers indicate the best results and underlined numbers indicate the second-best results within each model scale.}
\label{tab-privileged-ablation}

\newcommand{\hcell}[1]{\begin{tabular}[c]{@{}c@{}}#1\end{tabular}}

\resizebox{\textwidth}{!}{\begin{tabular}{c|c|cccccc|c|ccc|c}
\toprule
& & \multicolumn{7}{c|}{STEM Visual Reasoning Benchmarks} & \multicolumn{4}{c}{General Visual Reasoning Benchmarks} \\
\midrule
\hcell{Privileged\\Information} & \hcell{Teacher\\Score} & \hcell{MathVerse$_{\mathrm{m}}^{\mathrm{v}}$} & \hcell{MathVerse$_{\mathrm{m}}^{\mathrm{t}}$} & \hcell{DynaMath} & \hcell{WeMath} & \hcell{MathVista$_{\mathrm{m}}$} & \hcell{MathVision$_{\mathrm{m}}$} & \hcell{\textbf{\, Avg.\, }} & \hcell{MMMU$_{\mathrm{pro}}^{\mathrm{v}}$} & \hcell{MMMU$_{\mathrm{pro}}^{\mathrm{std}}$} & \hcell{HallBen} & \hcell{\textbf{\, Avg.\, }} \\
\midrule

Student 2B & 53.29 & 34.14 & 51.14 & \underline{53.05} & 30.48 & 62.70 & \textbf{29.28} & 43.46 & 27.92 & 35.38 & \underline{68.98} & 44.09 \\
Reasoning & 95.63 & 40.74 & 56.85 & 52.73 & 34.00 & 64.00 & 24.67 & 45.49 & 31.50 & 37.28 & 66.35 & 45.05 \\
Answer & 97.19 & \underline{42.89} & \underline{57.87} & 52.38 & \underline{35.05} & \underline{66.20} & \underline{28.95} & \underline{47.22} & \textbf{31.97} & 37.34 & 68.03 & \underline{45.78} \\
Caption & 66.76 & 42.13 & 55.84 & 52.46 & 32.38 & 65.10 & 27.30 & 45.87 & 30.46 & \underline{37.51} & 68.56 & 45.51 \\
\textbf{Visual Cue} & 72.68 & \textbf{45.05} & \textbf{58.12} & \textbf{54.13} & \textbf{38.19} & \textbf{67.00} & 26.32 & \textbf{48.14} & \underline{31.73} & \textbf{37.63} & \textbf{69.93} & \textbf{46.43} \\
\midrule

Student 4B & 57.89 & 30.58 & 53.93 & 64.51 & 49.05 & 73.20 & 37.50 & 51.46 & 30.58 & 35.72 & 70.66 & 45.65 \\
Reasoning & 95.94 & 55.71 & 69.42 & \underline{65.63} & 38.29 & 74.40 & 38.16 & 56.93 & \underline{45.61} & 46.99 & 72.13 & 54.91 \\
Answer & 97.87 & \underline{57.74} & 71.45 & 64.71 & \underline{51.33} & 75.50 & \underline{43.09} & \underline{60.63} & 45.20 & \underline{49.42} & \textbf{73.19} & 55.93 \\
Caption & 74.15 & 56.98 & \underline{71.70} & 65.05 & 51.24 & \underline{75.80} & 41.45 & 60.37 & \textbf{45.66} & \underline{49.42} & \underline{72.87} & \underline{55.98} \\
\textbf{Visual Cue} & 81.05 & \textbf{59.39} & \textbf{72.08} & \textbf{65.89} & \textbf{51.62} & \textbf{76.80} & \textbf{44.08} & \textbf{61.64} & 45.49 & \textbf{49.71} & \textbf{73.19} & \textbf{56.13} \\
\midrule

Student 8B & 60.31 & 36.55 & 67.26 & \underline{67.37} & 53.90 & 76.60 & \textbf{46.38} & 58.01 & 42.43 & 41.27 & 74.45 & 52.72 \\
Reasoning & 96.62 & 58.88 & 70.30 & 65.79 & 50.38 & \underline{76.70} & 39.80 & 60.31 & 47.51 & 46.82 & 73.50 & 55.95 \\
Answer & 97.87 & 55.58 & 68.40 & 67.07 & \underline{55.14} & 75.00 & 44.74 & 60.99 & 44.51 & 48.50 & \underline{74.55} & 55.85 \\
Caption & 75.01 & \underline{60.79} & \textbf{73.73} & 66.57 & 50.10 & 75.40 & 43.09 & \underline{61.61} & \underline{48.38} & \underline{51.21} & 74.13 & \underline{57.91} \\
\textbf{Visual Cue} & 82.32 & \textbf{61.55} & \underline{73.10} & \textbf{68.82} & \textbf{55.33} & \textbf{76.80} & \underline{45.07} & \textbf{63.44} & \textbf{52.49} & \textbf{53.76} & \textbf{74.87} & \textbf{60.37} \\
\bottomrule
\end{tabular}}
\end{table*}

\section{Conclusion}

In this work, we have proposed Visual Cue On-Policy Self-Distillation (VC-OPD), a framework that transfers the visual reasoning capabilities of large advanced MLLMs to smaller models through question-conditioned visual cues. By introducing a Visual Cue Agent and multi-stage quality filtering, VC-OPD constructs reliable privileged information that enables the teacher to provide more informative supervision on student-generated trajectories. The student can thus internalize the ability to identify and utilize critical visual evidence without requiring additional information during inference.
Extensive experiments on STEM and general visual reasoning benchmarks demonstrate that VC-OPD consistently improves the performance of MLLMs across different scales and outperforms existing reinforcement learning and distillation baselines. These results highlight the effectiveness of question-conditioned visual cues as transferable privileged information and provide a promising direction for building more efficient and capable small-scale multimodal models.

\bibliography{reference}
\bibliographystyle{conference}

\newpage

\appendix

\section{Training Dataset and Benchmarks}

\subsection{Training Dataset}
\label{sec-Training-Dataset}
\textbf{ViRL39K.}
We adopt ViRL39K~\citep{ViRL39K} as the training dataset. ViRL39K contains approximately 39K carefully curated multimodal reasoning samples, each accompanied by a verifiable reasoning process and answer, providing high-quality supervision signals for learning complex vision-language reasoning processes.
The dataset covers a wide range of domains, including STEM fields such as mathematics, physics, chemistry, and biology, as well as broader social science topics, enabling models to learn domain knowledge and reasoning abilities across diverse fields.
Furthermore, ViRL39K contains diverse visual inputs, including geometric figures, charts, tables, documents, and spatial relations, allowing models to learn visual understanding and cross-modal reasoning abilities across various visual scenarios.

\subsection{Evaluation Benchmarks}
\label{sec-Benchmarks}
To comprehensively evaluate the multimodal reasoning ability of our model, we select a set of representative benchmarks covering visual mathematical reasoning, general cross-disciplinary understanding, and visual reliability. All evaluations are conducted on publicly available test sets.

\textbf{MathVerse}
~\citep{mathverse} is a comprehensive benchmark designed to evaluate visual mathematical reasoning ability. It constructs multiple versions of the same mathematical problem with varying levels of visual and textual information, enabling analysis of whether models genuinely rely on visual content during reasoning.
We evaluate on its mini subset and report results under both Vision Only and Text Dominant settings, denoted as MathVerse\(_m^v\) and MathVerse\(_m^t\), respectively.

\textbf{DynaMath}
~\citep{dynamath} generates multiple variants of the same problem by dynamically modifying numerical values, function plots, and other visual conditions.
It is designed to evaluate the robustness and generalization ability of models under variations in visual mathematical tasks.
This benchmark primarily examines whether models can extract stable mathematical relationships from visual information rather than relying on fixed patterns or superficial cues.

\textbf{WeMath}
~\citep{Wemath} evaluates model capabilities from the perspectives of mathematical knowledge structures and multi-step problem decomposition.
It includes compositional reasoning tasks involving different mathematical concepts, aiming to measure models' understanding of fundamental mathematical knowledge and their ability to transfer knowledge during complex reasoning processes.

\textbf{MathVista}
~\citep{mathvista} is a comprehensive visual mathematical reasoning benchmark covering various task types, including arithmetic, algebra, geometry, statistics, science, and logical reasoning.
We evaluate our model on its mini test subset, denoted as MathVista\(_m\).

\textbf{MathVision}
~\citep{mathvision} collects high-quality visual mathematical problems from real-world mathematical competition settings, covering multiple mathematical domains and difficulty levels.
We evaluate on its mini subset, denoted as MathVision\(_m\), to further assess model performance in challenging visual mathematical reasoning scenarios.

\textbf{MMMU-Pro}
~\citep{Mmmu-pro} is an enhanced benchmark for evaluating general multimodal reasoning capabilities.
Compared with the original MMMU, MMMU-Pro filters out questions that can be answered solely based on textual information, introduces more challenging distractor options, and incorporates a Vision Only setting where question texts are embedded into images.
These designs provide a stricter evaluation of visual perception, text understanding, and cross-modal reasoning abilities. We report results under both the Vision Only and standard multimodal settings, denoted as MMMU\(_{pro}^v\) and MMMU\(_{pro}^{std}\), respectively.

\textbf{HallusionBench}
~\citep{hallusionbench} is designed to evaluate visual reliability in large vision-language models, with a focus on hallucination issues caused by misleading visual information or linguistic contexts.
It measures whether models can accurately understand visual content, avoid visual illusions, and mitigate language biases during multimodal reasoning. In the tables, HallusionBench is abbreviated as HallBen.

\section{Ablation Study on GRPO Teacher Warm-up}
\label{sec-exp-warm-up}
To investigate the role of GRPO teacher warm-up in VC-OPD, we consider four training configurations: the base model ({base}), OPD with the base model directly serving as the teacher ({base+OPD}), GRPO warm-up alone without subsequent distillation ({base+Warm Up}), and the full VC-OPD pipeline, where OPD is performed with the GRPO-warmed model as the teacher ({base+Warm Up+OPD}). We conduct this ablation across three model scales (2B, 4B, and 8B), with results summarized in Table~\ref{tab-GRPO-warm-up}.

\begin{table*}[h]
\centering
\caption{Ablation results of GRPO warm-up across three model scales: 2B, 4B, and 8B. Teacher Score denotes the score of the teacher model on the training set when visual cues are provided. For benchmark results, bold numbers indicate the best results, while underlined numbers indicate the second-best results within each model scale.}
\label{tab-GRPO-warm-up}
\setlength{\tabcolsep}{0.7pt}
\renewcommand{\arraystretch}{1.08}
\resizebox{\textwidth}{!}{\begin{tabular}{l|c|cccccc|c|ccc|c}
\toprule
\multirow{2}{*}{Methods}
& \multirow{2}{*}{\shortstack{Teacher \\ Score}}
& \multicolumn{7}{c|}{STEM Visual Reasoning Benchmarks}
& \multicolumn{4}{c}{General Visual Reasoning Benchmarks} \\
\cmidrule(lr){3-9}\cmidrule(lr){10-13}
&
&
MathVerse$_{\mathrm{m}}^{\mathrm{v}}$
& MathVerse$_{\mathrm{m}}^{\mathrm{t}}$
& DynaMath
& WeMath
& MathVista$_{\mathrm{m}}$
& MathVision$_{\mathrm{m}}$
& \textbf{\, Avg.\, }
& MMMU$_{\mathrm{pro}}^{\mathrm{v}}$
& MMMU$_{\mathrm{pro}}^{\mathrm{std}}$
& HallBen
& \textbf{\, Avg.\, } \\
\midrule

base 2B
& --
& 34.14
& 51.14
& 53.05
& 30.48
& \underline{62.70}
& \textbf{29.28}
& 43.46
& 27.92
& 35.38
& 68.98
& 44.09 \\

+OPD
& 64.18
& 33.50
& 49.37
& 50.96
& 27.90
& 61.60
& 26.64
& 41.66
& 27.28
& \underline{36.01}
& \underline{69.40}
& 44.23 \\

+Warm Up
& --
& \underline{35.66}
& \underline{53.68}
& \textbf{54.19}
& \underline{35.71}
& 62.40
& \underline{28.29}
& \underline{44.99}
& \underline{30.98}
& 35.43
& 67.19
& \underline{44.54} \\

\textbf{+Warm Up+OPD}
& 72.68
& \textbf{45.05}
& \textbf{58.12}
& \underline{54.13}
& \textbf{38.19}
& \textbf{67.00}
& 26.32
& \textbf{48.14}
& \textbf{31.73}
& \textbf{37.63}
& \textbf{69.93}
& \textbf{46.43} \\
\midrule

base 4B
& --
& 30.58
& 53.93
& 64.51
& 49.05
& 73.20
& 37.50
& 51.46
& 30.58
& 35.72
& 70.66
& 45.65 \\

+OPD
& 71.58
& \underline{52.03}
& \underline{68.40}
& \underline{65.39}
& 45.05
& \underline{75.70}
& 39.80
& \underline{57.73}
& \underline{43.01}
& \underline{48.27}
& \underline{71.08}
& \underline{54.12} \\

+Warm Up
& --
& 32.36
& 57.49
& 63.59
& \underline{50.29}
& 74.60
& \textbf{45.07}
& 53.90
& 30.12
& 35.49
& 64.35
& 43.32 \\

\textbf{+Warm Up+OPD}
& 81.05
& \textbf{59.39}
& \textbf{72.08}
& \textbf{65.89}
& \textbf{51.62}
& \textbf{76.80}
& \underline{44.08}
& \textbf{61.64}
& \textbf{45.49}
& \textbf{49.71}
& \textbf{73.19}
& \textbf{56.13} \\
\midrule

base 8B
& --
& 36.55
& 67.26
& 67.37
& 53.90
& \underline{76.60}
& \underline{46.38}
& 58.01
& 42.43
& 41.27
& \underline{74.45}
& 52.72 \\

+OPD
& 74.84
& \underline{56.35}
& \underline{70.94}
& \underline{67.50}
& 53.62
& 75.70
& \textbf{46.71}
& \underline{61.80}
& \underline{50.40}
& \underline{52.83}
& 74.24
& \underline{59.16} \\

+Warm Up
& --
& 37.44
& 66.24
& 67.01
& \underline{55.14}
& 76.00
& 43.09
& 57.49
& 37.51
& 40.17
& 72.98
& 50.22 \\

\textbf{+Warm Up+OPD}
& 82.32
& \textbf{61.55}
& \textbf{73.10}
& \textbf{68.82}
& \textbf{55.33}
& \textbf{76.80}
& 45.07
& \textbf{63.44}
& \textbf{52.49}
& \textbf{53.76}
& \textbf{74.87}
& \textbf{60.37} \\
\bottomrule
\end{tabular}}
\end{table*}

Across all three model scales, the full pipeline consistently achieves the best average performance on both STEM and general visual reasoning benchmarks. Compared with directly applying OPD without teacher warm-up, incorporating GRPO warm-up improves the STEM average by 6.48, 3.91, and 1.64 points for the 2B, 4B, and 8B models, respectively. Similar improvements are observed on general visual reasoning benchmarks, with gains of 2.20, 2.01, and 1.21 points across the three model scales. The full pipeline also substantially outperforms the corresponding base models. Taking the 4B model as an example, it improves the STEM and general averages by 10.18 and 10.48 points, respectively. These consistent gains across model scales and benchmark groups demonstrate the effectiveness of GRPO teacher warm-up in enhancing subsequent OPD.

GRPO warm-up consistently improves the teacher model's performance on the training set when privileged information is available. The Teacher Score improves by 8.50, 9.47, and 7.48 points for the 2B, 4B, and 8B models, respectively. This consistent improvement demonstrates that GRPO warm-up enables the teacher to better leverage privileged visual information and strengthens its reasoning capability under privileged supervision. The resulting stronger teacher provides higher-quality supervision during OPD, facilitating more effective transfer of privileged knowledge to the student.

\section{Display of performance improvement}
\label{sec-display-improve}

To further provide an intuitive analysis of how VC-OPD improves the visual reasoning capabilities of the model, we present representative examples from two perspectives: the privileged teacher and the student after distillation. The former is used to examine whether question-conditioned visual cues can directly compensate for deficiencies in visual information acquisition, while the latter demonstrates whether such capabilities can be internalized by the student through on-policy distillation and retained at inference time without access to additional visual cues.

\begin{figure}[t]
\centering
\centerline{\includegraphics[width=\linewidth]{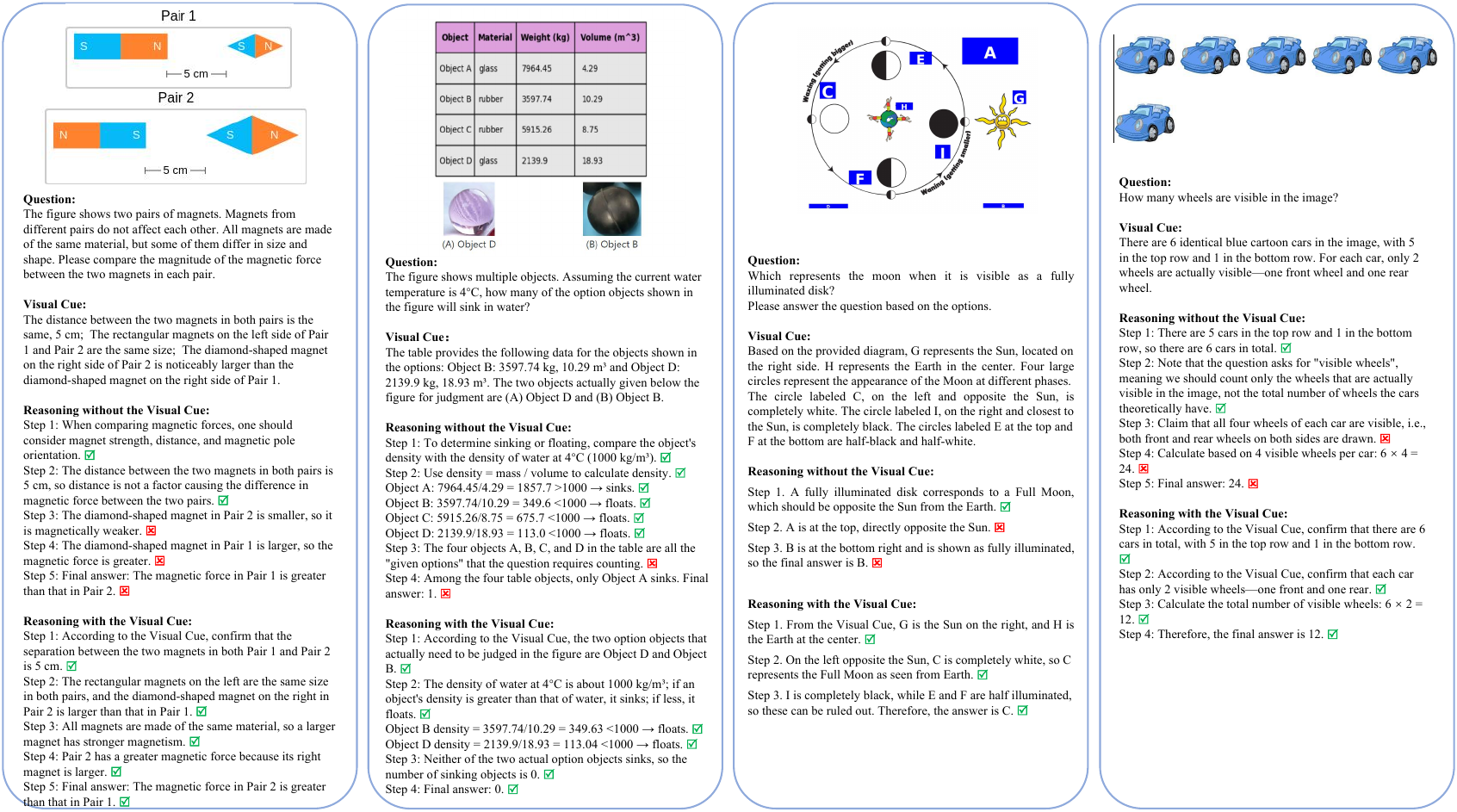}}
\caption{Qualitative examples showing how Visual Cues improve the privileged teacher's visual reasoning. Visual Cues provide critical question-conditioned evidence that corrects visual perception and grounding errors, enabling accurate subsequent reasoning.
}
\label{fig-teacher-cap}
\end{figure}
\begin{figure}[!h]
\centering
\centerline{\includegraphics[width=\linewidth]{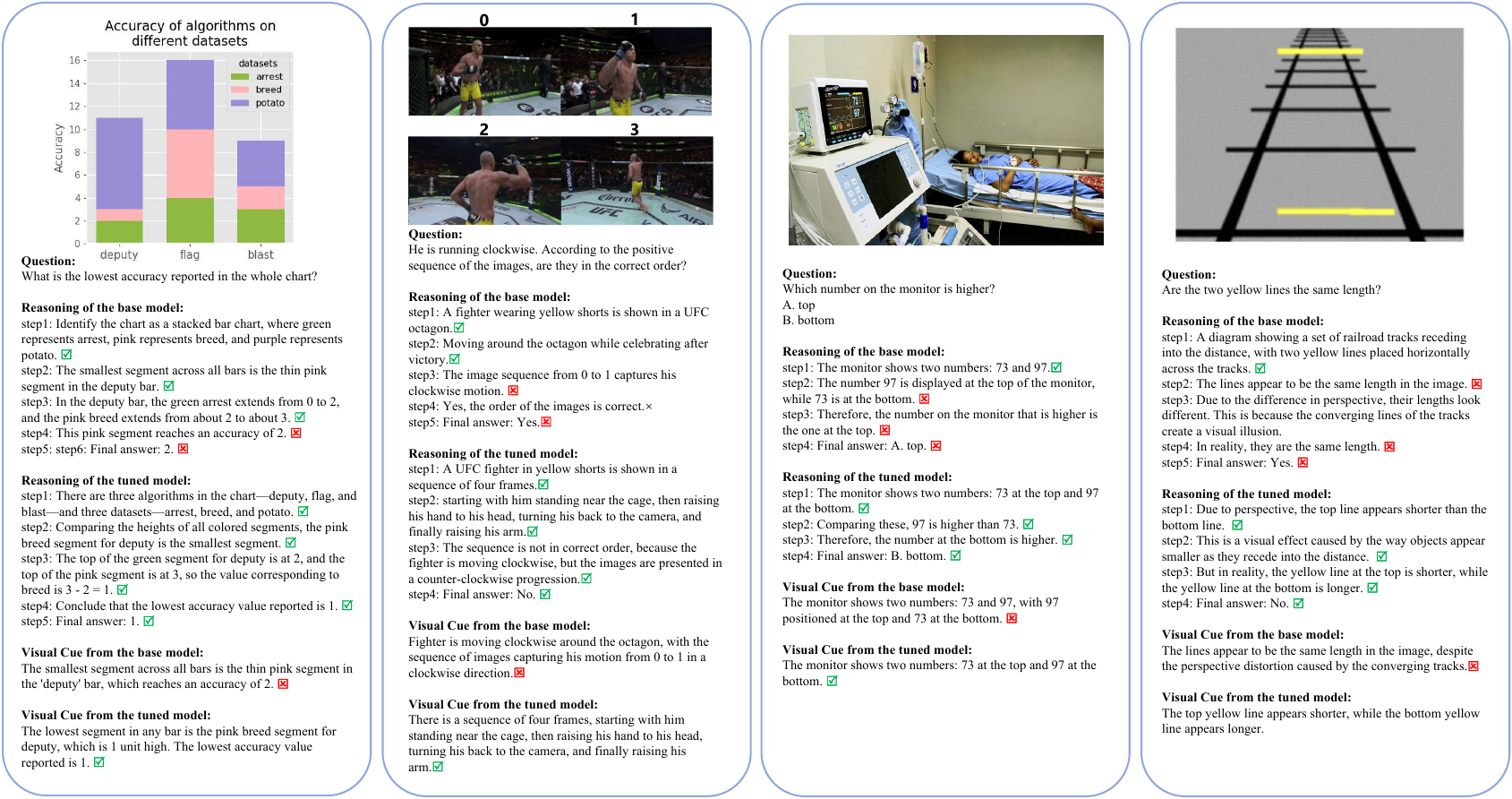}}
\caption{Qualitative examples showing the capability improvement of the student after VC-OPD. Without access to Visual Cues at inference time, the trained student better identifies and utilizes question-conditioned visual evidence directly from the original image.
}
\label{fig-student-cap}
\end{figure}

As shown in Fig.~\ref{fig-teacher-cap}, without the Visual Cue, the teacher can already apply the correct reasoning principles on some samples, yet still arrives at incorrect conclusions because it fails to accurately identify the key visual facts required to solve the problem. For example, the model may misjudge the relative sizes of different objects, incorrectly recognize labels and their spatial positions in the figure, miscount the number of actually visible objects, or confuse information in a table with the objects that the question specifically asks about. After incorporating the Visual Cue, the teacher gains direct access to question-conditioned visual evidence, allowing it to correct these perceptual or visual grounding errors and subsequently perform the reasoning correctly. This suggests that the primary role of the Visual Cue is not to directly provide the final answer, but rather to help the model accurately acquire the critical visual information required for subsequent reasoning.

Fig.~\ref{fig-student-cap} further illustrates how this capability is transferred to the student through VC-OPD. Unlike the privileged teacher, the trained student has no access to any Visual Cue at inference time and receives only the original image and question as input. Nevertheless, after VC-OPD training, the student can more accurately attend to question-conditioned visual regions and relationships, recover critical evidence that the base model tends to overlook or misinterpret, and thereby perform the subsequent reasoning correctly. This observation indicates that VC-OPD does not make the model dependent on external visual descriptions at inference time. Instead, through token-level supervision provided by the privileged teacher, it progressively internalizes into the student policy the capability to extract and utilize the visual information captured by the Visual Cue.

Taken together, the two groups of examples show that Visual Cues first induce more reliable visual understanding and reasoning behavior on the teacher-side, after which VC-OPD transfers this advantage to a student that operates without any additional privileged information. These qualitative results are consistent with the preceding quantitative findings and further suggest that the performance gains of VC-OPD are associated with improved identification and utilization of question-conditioned visual evidence.

\section{Hyperparameter settings}
\label{sec-Hyperparameter}
Table~\ref{tab-hyperparam} summarizes the main training and evaluation hyperparameters of VC-OPD. Unless otherwise specified, all experiments use the same configuration.
Experiments are implemented using PyTorch 2.8.0 and vLLM 0.18.0.

During the Visual Cue Agent stage, we perform $N_r=8$ rollouts for each sample and set the lower and upper thresholds of the Difficulty Gate to $\tau_d^{-}=1$ and $\tau_d^{+}=7$, respectively, thereby filtering out samples that are either nearly unsolvable or already solved reliably by the student model. For the Reliability Gate, we set $\tau_{\mathrm{rel}}=7$, retaining only visual cues that enable the teacher to answer correctly in at least 7 out of 8 rollouts. 

For VC-OPD training, we use a batch size of 96 and sample 8 responses per prompt. The maximum prompt and response lengths are set to 8192 and 4096 tokens, respectively, and all models are trained for one epoch. Token-level JSD distillation is performed over the top-100 vocabulary tokens selected by the student distribution, with the JSD interpolation coefficient set to $\alpha=0.5$. The teacher remains frozen throughout distillation, and no additional reward model is used.

We use the AdamW optimizer with a learning rate of $2\times10^{-6}$ and a constant learning-rate scheduler. All evaluations are conducted using VLMEvalKit, with vLLM as the inference engine and a maximum generation length of 4096 tokens. For benchmarks requiring model-based evaluation, we use Qwen3.5-27B as the judge model with deterministic decoding, setting temperature $=0$ and top-$p=1.0$.

\begin{table*}[t]
\centering
\caption{Hyperparameter settings.}
\label{tab-hyperparam}
\setlength{\tabcolsep}{10pt}
\renewcommand{\arraystretch}{1.08}
\begin{tabular}{c c c}
\toprule
\textbf{Category} & \textbf{Hyperparameter} & \textbf{Value} \\
\midrule
Visual Cue Agent & Rollout ($N_r$) & 8 \\
Visual Cue Agent & Lower threshold of Difficulty Gate $\tau_d^{-}$ & 1 \\
Visual Cue Agent & Upper threshold of Difficulty Gate $\tau_d^{+}$ & 7 \\
Visual Cue Agent & Threshold of Reliability Gate $\tau_{\mathrm{rel}}$ & 7 \\
\midrule
Data & Train batch size & 96 \\
Data & PPO mini-batch size & 96 \\
Data & Samples per prompt & 8 \\
Data & Max prompt length & 8192 \\
Data & Max response length & 4096 \\
Data & Total training epochs & 1 \\
\midrule
Distillation & Distillation top-$k$ & 100 \\
Distillation & Distillation top-$k$ source & student distribution \\
Distillation & JSD interpolation coefficient ($\alpha$) & 0.5 \\
Distillation & Teacher model update & Frozen \\
Distillation & Reward model & Disabled \\
\midrule
Optimization & Learning rate & $2\times10^{-6}$ \\
Optimization & LR scheduler type & constant \\
Optimization & Optimizer & AdamW \\
Optimization & Gradient checkpointing & True \\
Optimization & Dynamic batch size & True \\
\midrule
Evaluation & Evaluation Tool & VLMEvalKit \\
Evaluation & Inference engine & vLLM \\
Evaluation & Model class & LMDeployAPI \\
Evaluation & Max generation tokens & 4096 \\
Evaluation & Judge model & Qwen3.5-27B \\
Evaluation & Sampling temperature & 0.0 \\
Evaluation & Top-$p$ & 1.0 \\
\bottomrule
\end{tabular}
\end{table*}

\section{LLM-Assisted Language Polishing}

Large language models (LLMs) were used solely for language polishing, including improving grammar, clarity, fluency, and readability of the manuscript. LLMs were not used for research ideation, methodology design, experimental design or implementation, data analysis, interpretation of results, or formulation of scientific claims and conclusions. All LLM-assisted edits were reviewed and, where necessary, revised by the authors. The authors take full responsibility for the final content of the manuscript.

\end{document}